\documentclass[11pt,a4paper]{article}
\usepackage[hyperref]{acl2020}
\usepackage{times}
\usepackage{latexsym}
\usepackage{graphicx}
\usepackage[labelformat=simple]{subcaption}

\usepackage{tabu}
\usepackage{multirow}
\usepackage{url}
\usepackage{amsmath}
\usepackage{amssymb}
\usepackage{xr}

\aclfinalcopy 
\def\aclpaperid{664} 

\title{Designing Precise and Robust Dialogue Response Evaluators}

\author{Tianyu Zhao \\\And
Divesh Lala \\
Graduate School of Informatics \\
Kyoto University \\
{\tt \{zhao,lala,kawahara\}@sap.ist.i.kyoto-u.ac.jp} \\\And
Tatsuya Kawahara \\
}

\date{}

\begin{document}
\maketitle

\begin{abstract}
Automatic dialogue response evaluator has been proposed as an alternative to automated metrics and human evaluation. However, existing automatic evaluators achieve only moderate correlation with human judgement and they are not robust. In this work, we propose to build a reference-free evaluator and exploit the power of semi-supervised training and pretrained (masked) language models. Experimental results demonstrate that the proposed evaluator achieves a strong correlation ($>$ 0.6) with human judgement and generalizes robustly to diverse responses and corpora. We open-source the code and data in \url{https://github.com/ZHAOTING/dialog-processing}. 
\end{abstract}

\section{Introduction}
\label{sec:intro}

Evaluation of conversational systems has been one major obstacle in dialogue research. Particularly for open-domain dialogues, automated metrics have been shown to correlate poorly with human judgement~\citep{liu2016not}. Although human evaluation provides the most accurate assessment, they are slow and expensive. An alternative is to train an evaluator that learns to predict a human-like score. \citet{lowe2017towards} proposed ADEM, a supervised regression model, for automatic response evaluation and reported 0.436 Pearson's and 0.428 Spearman's correlations with human judgement. Though better than automated metrics, the scores only indicate moderate correlations. Another criticism from~\citet{sai2019reeval} further pointed out that ADEM produces scores of low deviation and lacks robustness under adversarial attack. 

An ideal evaluator should be precise such that its predictions have a strong correlation with human judgement. It should also be robust such that it generalizes to new dialogues unseen during training. We explored three methods to improve the precision and robustness of response evaluators. 1) We propose building reference-free evaluator since reference-dependent metrics cause the problem of low deviation described by \citet{sai2019reeval}. We also find that the reference-dependent evaluators' performance degrades significantly when we remove ground-truth responses from test data. 2) \citet{tao2018ruber} proposed an unsupervised model (RUBER) that outperforms supervised ADEM by training on a next sentence prediction (NSP) task. We show that RUBER can be further improved by supervised training on a small amount of annotated data. 3) We make use of strong pretrained models such as RoBERTa~\citep{liu2019roberta} to obtain better text representations. By combining the three methods, a reference-free, semi-supervised, RoBERTa-based evaluator has better correlation and robustness. Experimental results also show that the model can maintain good performances in cross-domain and low-resource settings.

\section{Related Works}

\textbf{Automatic response evaluator} was first proposed by \citet{lowe2017towards} to mimic human annotator's assessment of response appropriateness. They collected human annotations of response quality for 4,104 context-response pairs, and train a regression network (ADEM) supervisedly by minimizing a squared error. \citet{tao2018ruber} proposed an unsupervised method (RUBER) to train automatic evaluators, where a model is optimized to distinguish a ground-truth response and a negative-sampling response by minimizing a margin rank loss. This process resembles the next sentence prediction (NSP) task applied in the training of BERT~\citep{devlin2019bert}. It allows for exploiting a large amount of conversation data and has been shown to outperform ADEM. Using ADEM and RUBER as the baselines of this work, we will analyze their shortcomings and develop solutions to build more precise and robust evaluators.

\textbf{Next sentence prediction} is to predict whether a sentence is a true continuation given a preceding context, where a positive sample is the ground-truth subsequent sentence and a negative sample is a different piece of text. NSP benefits not only evaluation~\citep{tao2018ruber}, but also language understanding~\citep{devlin2019bert} and language generation~\citep{bruni2017adversarial,wolf2019transfertransfo}.

\textbf{Dialogue response evaluation} can also be improved with better automated metrics and approximation to response quality. Examples of successful attempts to improve automated metrics include exploiting multiple references for comparison~\citep{gupta2019investigating} and combining human judgement with automated metrics~\citep{hashimoto2019unifying}. \citet{li2019acute} demonstrated that single-turn human judgement is not reliable as expected and proposed multi-turn human evaluation. \citet{ghandeharioun2019approximating} approximated sentiment, semantic similarity, and engagement with new automated metrics and used a hybrid metric in a multi-turn evaluation setting. \citet{dziri2019evaluating} showed that entailment is also an option to approximate dialogue coherence and quality.

\begin{figure*}[t]
    \centering
    \begin{subfigure}{.25\textwidth}
        \centering
        \includegraphics[width=1.0\linewidth]{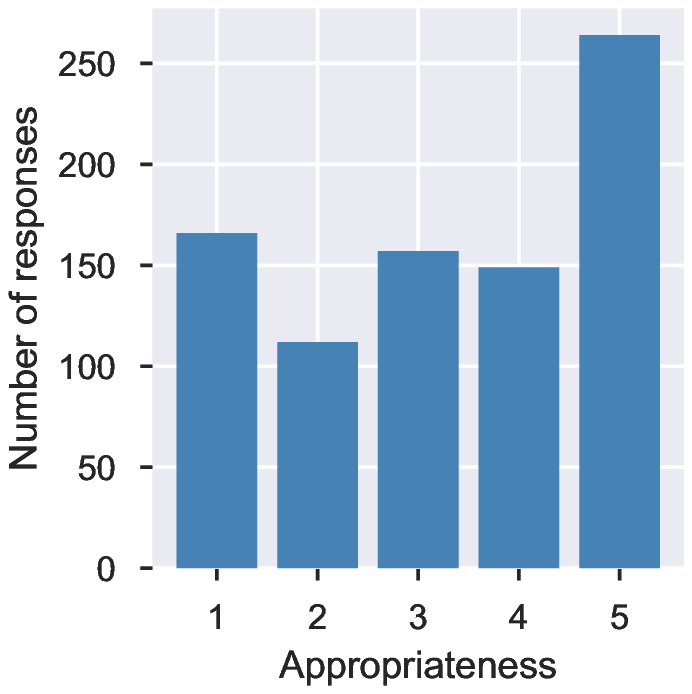}  
        \caption{Overall score distribution.}
        \label{fig:anno_score_dist}
    \end{subfigure}
    \begin{subfigure}{.6\textwidth}
        \centering
        \includegraphics[width=1.0\linewidth]{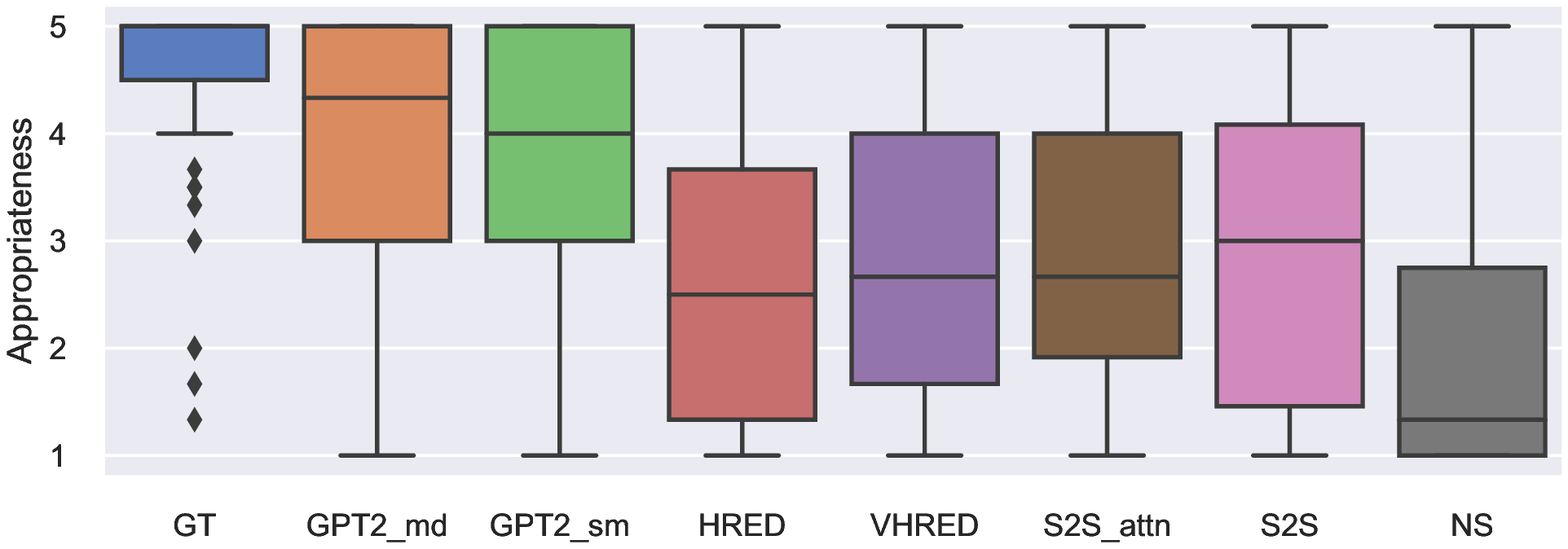}  
        \caption{Box plot of scores for each response source. GT - ground-truth, NS - negative-sampling.}
        \label{fig:anno_sys_box}
    \end{subfigure}
\caption{Distributions of human annotations on response \textit{appropriateness}~(\S\ref{sec:data}). }
\label{fig:anno_analysis}
\end{figure*}

\section{Background}

ADEM is a regression model that takes as inputs a dialogue context vector $\mathbf{c}$, a hypothesis response vector $\mathbf{\hat{r}}$, and a reference response vector $\mathbf{r}$. Its output is the sum of a referenced metric and an unreferenced metric:
\begin{align}
    \text{ADEM}_\textit{ref}(\mathbf{r}, \mathbf{\hat{r}}) &= \mathbf{r}^T N \mathbf{\hat{r}}, \\
    \text{ADEM}_\textit{unref}(\mathbf{c}, \mathbf{\hat{r}}) &= \mathbf{c}^T M \mathbf{\hat{r}}, \label{equ:adem_unref}
\end{align}
where the encoding vectors are produced by pretrained RNN encoders. $M$ and $N$ are trainable parameters. 

RUBER also combines two metrics but computes them differently:
\begin{align}
    \text{RUBER}_\textit{ref}(\mathbf{r}, \mathbf{\hat{r}}) &= \frac{\mathbf{r}^T \mathbf{\hat{r}}}{\lVert \mathbf{r} \rVert \cdot \lVert \mathbf{\hat{r}} \rVert }, \\
    \text{RUBER}_\textit{unref}(\mathbf{c}, \mathbf{\hat{r}}) &= \text{MLP}([\mathbf{c};\mathbf{\hat{r}};\mathbf{c}^T M \mathbf{\hat{r}}]; \theta),
\end{align}
where $[\cdot;\cdot]$ denotes the concatenation of vectors and $\text{MLP}$ is a multi-layer perceptron with nonlinear activation functions. $M$ and $\theta$ are trainable parameters.

Besides the differences in metric computation, they are different in training strategy. ADEM uses supervised training to minimize the mean square error between predictions and human scores, while RUBER uses unsupervised training on an NSP task to minimize a margin ranking loss. In Section~\ref{sec:method}, we combine their advantages to build a better response evaluator.

\begin{table*}
    \centering
    \begin{tabu} to 1.0\columnwidth {l|cc|c|cc|c}
        \multirow{3}{*}{\textbf{Model}} & \multicolumn{3}{c|}{\textbf{Full Test Data}} & \multicolumn{3}{c}{\textbf{Excluding Ground-truth}} \\
        & \multicolumn{3}{c|}{\textbf{(90 responses)}} & \multicolumn{3}{c}{\textbf{(77 responses)}} \\
        \cline{2-7}
        & \textbf{Pearson} & \textbf{Spearson} & \textbf{SD} & \textbf{Pearson} & \textbf{Spearson} & \textbf{SD} \\
        \hline
        \hline
        ADEM & & & & & & \\
        \quad \textit{full} & 0.34$^{**}$ & 0.36$^{**}$ & 0.51 & 0.25 & 0.23 & 0.30 \\
        \quad \textit{ref.} & 0.32$^{*}$ & 0.35$^{**}$ & 0.52 & 0.21 & 0.23 & 0.30 \\
        \quad \textit{unref.} & 0.26 & 0.26 & 0.32 & 0.28 & 0.27 & 0.33 \\
        \hline
        RUBER & & & & & & \\
        \quad \textit{full} & 0.37$^{**}$ & 0.31$^{*}$ & 0.67 & 0.43$^{**}$ & 0.41$^{**}$ & 0.68 \\
        \quad \textit{ref.} & 0.32$^{*}$ & 0.29$^{*}$ & 0.07 & 0.12 & 0.13 & 0.04 \\
        \quad \textit{unref.} & 0.35$^{**}$ & 0.29$^{*}$ & 1.32 & 0.43$^{**}$ & 0.39$^{**}$ & 1.35 \\
        \hline
        Human & 1.0 & 1.0 & 1.42 & 1.0 & 1.0 & 1.40 
    \end{tabu}
    \caption{Comparison between referenced metric and unreferenced metric on the full test data and the ground-truth response-excluded test data~(\S\ref{ssec:ref_free}). \textbf{SD} is short for standard deviation. $^{*}$~denotes scores that have p-values~$<$~0.01. $^{**}$~denotes scores that have p-values~$<$~0.001. }
    \label{tab:ref_res}
\end{table*}

\section{Data Collection}
\label{sec:data}

For assessing dialogue response evaluators, we sample 100 dialogues from the test split of the DailyDialog corpus~\cite{li2017dailydialog} which contains 13,118 open-domain and human-written conversations. We expand them with extra response hypotheses and collect human annotations of response quality.

\textbf{Collection of Extra Responses.} Besides the ground-truth response, we add responses from different sources for each dialogue context, including 1) a negative-sampling response randomly selected from a different dialogue and 2) responses generated by generative models trained on the training split. We combine 6 generative models (S2S~\citep{sutskever2014sequence}, attentional S2S, HRED~\citep{serban2016building}, VHRED~\citep{serban2017hierarchical}, GPT2\textit{-sm}, and GPT2\textit{-md}~\citep{wolf2019transfertransfo}) with 3 decoding methods (greedy decoding, ancestral sampling, and nucleus sampling~\citep{holtzman2019curious}). The resulting response pool for each dialogue context contains 20 responses of various qualities.

\textbf{Collection of Human Annotations.} From the 2,000 dialogue-response pairs, we select 900 of them and ask Amazon Mechanical Turk workers to rate response \textit{appropriateness} on a 5-point Likert scale. Each pair is rated by four workers. After removing annotation outliers for each pair~\citep{leys2013detecting}, the remaining data reaches good reliability regarding an inter-annotator agreement with Krippendorff's $\alpha>0.8$~\citep{krippendorff2018content}.\footnote{More details of inter-annotator agreement and outlier removal are provided in Appendix~\ref{sec:agreement}.} We make a 0.8:0.1:0.1 split of the annotated data for training, validation and test.

Figure~\ref{fig:anno_score_dist} shows the overall distribution of 900 human scores on response \textit{appropriateness}, and Figure~\ref{fig:anno_sys_box} shows box plots of human scores for different response sources. The distributions suggest that the created data consists of diverse responses.

\section{Methodology}
\label{sec:method}
\subsection{Reference-free Evaluation}
\label{ssec:ref_free}
\citet{sai2019reeval} proved theoretically that the comparison with reference response in the referenced metric causes ADEM to make conservative predictions where scores have a very low standard deviation. To investigate the effect of removing reference from computation, we experiment with the full ADEM and RUBER as well as their referenced and unreferenced versions. As shown in Table~\ref{tab:ref_res}, the referenced metrics of ADEM and RUBER have much lower standard deviations than human scores. ADEM's unreferenced metric has low scores in both correlation and standard deviation because the full ADEM model is heavily affected by its referenced metric while its unreferenced metric is not fully utilized, especially in the data set that includes ground-truth responses. 

Another important finding is that the referenced metrics' correlations degrade significantly when we remove ground-truth responses from the test data. It suggests that referenced metrics may help evaluators to distinguish a ground-truth response from a non-ground-truth response easily, but they cannot distinguish a good response from a bad one among non-ground-truth responses. 

Based on the results, we propose to build reference-free evaluators and avoid direct comparison with reference responses to improve its robustness and diversity.

\subsection{Semi-supervised Training}
\label{ssec:semi_sup}
ADEM is a supervised model that relies on human annotations. However, it is expensive to collect large-scale annotated data; On the other hand, RUBER has been shown to reach reasonable correlation scores via only unsupervised training on an NSP task. A natural idea is to apply unsupervised training first and then finetune an evaluator using a relatively small amount of annotated data. Taking RUBER as an example, by finetuning RUBER on 720 annotated samples, we improve its Pearson's correlation from 0.37 to 0.45 and Spearman's correlation from 0.31 to 0.41. 

\begin{table}[!t]
    \centering
    \begin{tabu} to 1.0\columnwidth {l|cc|c}
        \multicolumn{1}{c|}{\textbf{Model}} & \textbf{Pr.} & \textbf{Spr.} & \textbf{Training data} \\
        \hline
        \hline
        RUBER & & & \\
        \enskip $\textit{sup.}$ & 0.37$^{**}$ & 0.31$^{*}$ & 130k \\
        \enskip $\textit{semi-sup.}$ & 0.45$^{**}$ & 0.41$^{**}$ & 130k+720 \\
    \end{tabu}
    \caption{Comparison between original unsupervised RUBER and semi-supervised RUBER~(\S\ref{ssec:semi_sup}). \textbf{Pr.} and \textbf{Spr.} are short for Pearson's correlation and Spearman's correlation, respectively. }
    \label{tab:ref_semisup}
\end{table}

\subsection{Powerful Text Encoder}
\label{ssec:encoder}
All the metrics mentioned before are based on encoding vectors $\mathbf{r}$, $\mathbf{\hat{r}}$ and $\mathbf{c}$, so a powerful text encoder is essential to building a good evaluator. ADEM and RUBER are both initialized with pretrained RNN response generators. As an alternative, pretrained (masked) language models such as BERT~\citep{devlin2019bert} and RoBERTa~\citep{liu2019roberta} can be used as a powerful text encoder and have benefited most downstream tasks in natural language processing~\citep{huang2019cosmos,lan2019albert,joshi2019spanbert,shimanaka2019machine}. We choose RoBERTa\textit{-large} to build our response evaluator.

A RoBERTa evaluator produces an encoding vector $\mathbf{d}$ given a context $c$ and a response $\hat{r}$ and then finally calculates its score via an MLP with a sigmoid function. We rescale the score to match annotator's scale of [1, 5]:
\begin{gather}
    \mathbf{d} = \text{RoBERTa}([c;\hat{r}]; \phi), \\
    \text{RoBERTa-eval}(c, \hat{r}) = 4 \cdot \text{MLP}(\mathbf{d}; \theta) + 1,
\end{gather}
where RoBERTa's parameter $\phi$ and MLP's parameter $\theta$ can both be optimized during training.

\section{Experimental Evaluations}
\label{sec:exp}
Table~\ref{tab:summary} shows the correlation scores and standard deviations of four metric groups. The first group is automated metrics that are based on $n$-gram overlapping (BLEU-2) or word embedding similarities (Average, Extrema, and Greedy). The second group is the baseline ADEM and RUBER. The third group is the semi-supervised full RUBER model, the semi-supervised unreferenced RUBER model, and the RoBERTa-based evaluator that combines the three proposed methods. Human scores are given in the final group. Semi-supervised training yields improvement in correlations, and abandoning referenced metrics makes predictions less conservative. The RoBERTa evaluator outperforms the baselines by a large margin and has a much human-like score diversity.

\begin{table}[!t]
    \centering
    \begin{tabu} to 1.0\columnwidth {l|cc|c}
        \multicolumn{1}{c|}{\textbf{Model}} & \textbf{Pr.} & \textbf{Spr.} & \textbf{SD} \\
        \hline
        \hline
        \multicolumn{4}{c}{Automated Metrics} \\
        \hline
        BLEU-2 & 0.31 & 0.23 & 0.31\\
        Average & 0.25 & 0.23 & 0.19 \\
        Extrema & 0.26 & 0.26 & 0.23 \\
        Greedy & 0.25 & 0.23 & 0.21 \\
        \hline
        \multicolumn{4}{c}{Baseline Evaluator} \\
        \hline
        ADEM & 0.34$^{**}$ & 0.36$^{**}$ & 0.51 \\
        RUBER & 0.37$^{**}$ & 0.31$^{*}$ & 0.67 \\
        \hline
        \multicolumn{4}{c}{Proposed Evaluator} \\
        \hline
        RUBER & & \\
        \quad $\textit{semi-sup.}$ & 0.45$^{**}$ & 0.41$^{**}$ & 0.42 \\
        \quad $\textit{unref.+semi-sup.}$ & 0.43$^{**}$ & 0.39$^{**}$ & 0.83 \\
        RoBERTa-eval & 0.64$^{**}$ & 0.66$^{**}$ & 1.26 \\
        \hline
        \multicolumn{4}{c}{Human Judgement} \\
        \hline
        Human & 1.0 & 1.0 & 1.42
    \end{tabu}
    \caption{Performances of automated metrics, baseline evaluators, and proposed evaluators~(\S\ref{sec:exp}). }
    \label{tab:summary}
\end{table}

\subsection{Transferability Study}
\label{ssec:transfer}
We are interested in applying a trained response evaluator to new data of different domains or styles. Therefore, we carry out experiments to study the transferability of the RoBERTa evaluator. In addition to the DailyDialog (DD) corpus, we further collect annotations on 900 responses from the PersonaChat (PC) corpus~\citep{zhang2018personalizing} following the same procedure in Section~\ref{sec:data}. The evaluator turns out to generalize to a new corpus much better than the baseline RUBER according to results in Table~\ref{tab:res_transfer}. The evaluator trained on the DD corpus achieves even higher correlation scores when applied to the PC corpus. However, performance degradation is observed when applying the evaluator trained on the PC corpus to the DD corpus. It suggests that we should make a careful choice of training data when planning to evaluate our models on different corpora.

\begin{table}[!t]
    \centering
    \begin{tabu} to 1.0\columnwidth {cc|cc}
        \multicolumn{2}{c|}{\textbf{Corpus}} & \multicolumn{2}{c}{\textbf{Correlation}} \\
        \hline
        \textbf{Train} & \textbf{Test} & \textbf{Pr.} & \textbf{Spr.} \\
        \hline
        \hline
        \multicolumn{4}{c}{RoBERTa evaluator} \\
        \hline
        DD & DD & 0.64$^{**}$ & 0.66$^{**}$ \\
        \hline
        DD & PC & 0.69$^{**}$ & 0.69$^{**}$ \\
        \hline
        PC & PC & 0.75$^{**}$ & 0.76$^{**}$ \\
        \hline
        PC & DD & 0.50$^{**}$ & 0.47$^{**}$ \\
        \hline
        \multicolumn{4}{c}{RUBER} \\
        \hline
        DD & DD & 0.37$^{**}$ & 0.31$^{*}$ \\
        \hline
        DD & PC & 0.12 & 0.17 \\
        \hline
        PC & PC & 0.58$^{**}$ & 0.57$^{**}$ \\
        \hline
        PC & DD & 0.06 & 0.06 \\
    \end{tabu}
    \caption{Correlations of RoBERTa evaluator and RUBER using training and test data from different corpora~(\S\ref{ssec:transfer}). }
    \label{tab:res_transfer}
\end{table}

\subsection{Low Resource Study}
\label{ssec:low_resource}
\begin{figure}[!t]
    \centering
    \includegraphics[width=1.0\columnwidth]{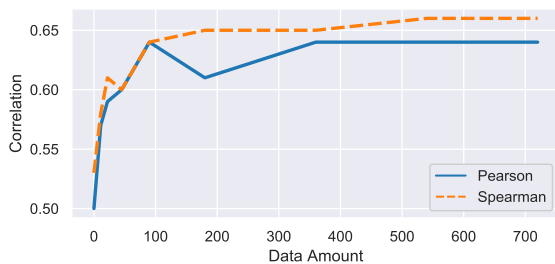}
    \caption{Performance of the RoBERTa evaluator w.r.t amount of supervised training data~(\S\ref{ssec:low_resource}). }
    \label{fig:res_data_amount}
\end{figure}

Although only 720 annotated samples are used in the experiments above, we explored the possibility of training with even fewer data. Figure~\ref{fig:res_data_amount} shows that, with only around 100 samples, the RoBERTa evaluator can reach performance close to the result obtained using the entire 720 samples.

\subsection{Robustness Evaluation}
\label{ssec:robust}
In this section, we address \citet{sai2019reeval}'s requirements towards a robust evaluator.

\textbf{1. Not be heavily influenced by the reference response.} The proposed evaluator is entirely independent of references.  

\textbf{2. Generalizing to diverse responses.} 1) After removing ground-truth from the test data, the RoBERTa evaluator still achieves 0.62 Pearson's correlation and 0.64 Spearman's correlation. 2) The evaluator achieves good performances on diverse responses~(see \S\ref{sec:data}) and different corpora~(see \S\ref{ssec:transfer}).


\textbf{3. Sensitivity to grammar and relevance of the response.} We also collected annotations for \textit{relevance} and \textit{grammatical correctness}. The RoBERTa evaluator trained on \textit{appropriateness} annotations can achieve 0.68 Pearson's and 0.67 Spearman's correlations with \textit{relevance} annotations, while its correlation scores with \textit{grammatical correctness} are only 0.09 and 0.15. However it is understandable because responses of perfect grammar can still be inappropriate in a certain context and grammar itself is not highly correlated with \textit{appropriateness}.\footnote{According to the collected annotations, \textit{appropriateness} and \textit{relevance} are highly correlated with 0.91 Pearson's and 0.91 Spearman's scores, while \textit{appropriateness} and \textit{grammatical correctness} have only 0.37 Pearson's and 0.34 Spearman's scores.}

\textbf{4. Robust against fooling attacks.} Unlike in~\citet{sai2019reeval}, we have not found any magic responses that can fool the evaluators to output high scores constantly.

\section{Conclusion}

Automatic dialogue response evaluators have problems in robustness and correlation with human judgement. We investigated three methods to alleviate them: 1) using reference-free metrics, 2) applying semi-supervised training, and 3) exploiting powerful pretrained text encoders. Experimental results demonstrated that our proposed evaluator achieved strong correlation ($>$ 0.6) with human judgement and showed robustness in dealing with diverse responses and a new domain. It can also be trained efficiently with less than 100 annotated samples. 

\section*{Acknowledgments}

The authors would like to thank Shinsuke Mori from Kyoto University, Wei Wu from Microsoft, Graham Neubig from CMU, and the anonymous reviewers for their constructive comments. This work was supported by JST ERATO Ishiguro Symbiotic Human-Robot Interaction program (Grant Number JPMJER1401), Japan.

\bibliography{../collection}

\begin{thebibliography}{26}
\expandafter\ifx\csname natexlab\endcsname\relax\def\natexlab#1{#1}\fi

\bibitem[{Bruni and Fernandez(2017)}]{bruni2017adversarial}
Elia Bruni and Raquel Fernandez. 2017.
\newblock Adversarial evaluation for open-domain dialogue generation.
\newblock In \emph{{SIGDIAL} 2017, The 18th Annual Meeting of the Special
  Interest Group on Discourse and Dialogue}, pages 284--288.

\bibitem[{Devlin et~al.(2019)Devlin, Chang, Lee, and
  Toutanova}]{devlin2019bert}
Jacob Devlin, Ming-Wei Chang, Kenton Lee, and Kristina Toutanova. 2019.
\newblock Bert: Pre-training of deep bidirectional transformers for language
  understanding.
\newblock In \emph{{NAACL} {HLT} 2019, The 2019 Conference of the North
  American Chapter of the Association for Computational Linguistics: Human
  Language Technologies}, pages 4171--4186.

\bibitem[{Dziri et~al.(2019)Dziri, Kamalloo, Mathewson, and
  Zaiane}]{dziri2019evaluating}
Nouha Dziri, Ehsan Kamalloo, Kory Mathewson, and Osmar~R Zaiane. 2019.
\newblock Evaluating coherence in dialogue systems using entailment.
\newblock In \emph{{NAACL} {HLT} 2019, The 2019 Conference of the North
  American Chapter of the Association for Computational Linguistics: Human
  Language Technologies}, pages 3806--3812.

\bibitem[{Ghandeharioun et~al.(2019)Ghandeharioun, Shen, Jaques, Ferguson,
  Jones, Lapedriza, and Picard}]{ghandeharioun2019approximating}
Asma Ghandeharioun, Judy~Hanwen Shen, Natasha Jaques, Craig Ferguson, Noah
  Jones, Agata Lapedriza, and Rosalind Picard. 2019.
\newblock Approximating interactive human evaluation with self-play for
  open-domain dialog systems.
\newblock In \emph{{NeurIPS} 2019, Advances in Neural Information Processing
  Systems 32}, pages 13658--13669.

\bibitem[{Gupta et~al.(2019)Gupta, Mehri, Zhao, Pavel, Eskenazi, and
  Bigham}]{gupta2019investigating}
Prakhar Gupta, Shikib Mehri, Tiancheng Zhao, Amy Pavel, Maxine Eskenazi, and
  Jeffrey~P Bigham. 2019.
\newblock Investigating evaluation of open-domain dialogue systems with human
  generated multiple references.
\newblock In \emph{{SIGDIAL} 2019 Workshop, The 20th Annual Meeting of the
  Special Interest Group on Discourse and Dialogue}, pages 379--391.

\bibitem[{Hashimoto et~al.(2019)Hashimoto, Zhang, and
  Liang}]{hashimoto2019unifying}
Tatsunori Hashimoto, Hugh Zhang, and Percy Liang. 2019.
\newblock Unifying human and statistical evaluation for natural language
  generation.
\newblock In \emph{{NAACL} {HLT} 2019, The 2019 Conference of the North
  American Chapter of the Association for Computational Linguistics: Human
  Language Technologies}, pages 1689--1701.

\bibitem[{Holtzman et~al.(2019)Holtzman, Buys, Forbes, and
  Choi}]{holtzman2019curious}
Ari Holtzman, Jan Buys, Maxwell Forbes, and Yejin Choi. 2019.
\newblock The curious case of neural text degeneration.
\newblock In \emph{{ICLR} 2020, The 5th International Conference on Learning
  Representations}.

\bibitem[{Huang et~al.(2019)Huang, Le~Bras, Bhagavatula, and
  Choi}]{huang2019cosmos}
Lifu Huang, Ronan Le~Bras, Chandra Bhagavatula, and Yejin Choi. 2019.
\newblock Cosmos qa: Machine reading comprehension with contextual commonsense
  reasoning.
\newblock In \emph{{EMNLP-IJCNLP} 2019, The 2019 Conference on Empirical
  Methods in Natural Language Processing and the 9th International Joint
  Conference on Natural Language Processing}, pages 2391--2401.

\bibitem[{Joshi et~al.(2020)Joshi, Chen, Liu, Weld, Zettlemoyer, and
  Levy}]{joshi2019spanbert}
Mandar Joshi, Danqi Chen, Yinhan Liu, Daniel~S Weld, Luke Zettlemoyer, and Omer
  Levy. 2020.
\newblock Spanbert: Improving pre-training by representing and predicting
  spans.
\newblock \emph{Transactions of the Association for Computational Linguistics},
  8:64--77.

\bibitem[{Krippendorff(2018)}]{krippendorff2018content}
Klaus Krippendorff. 2018.
\newblock \emph{Content analysis: An introduction to its methodology}.
\newblock Sage publications.

\bibitem[{Lan et~al.(2020)Lan, Chen, Goodman, Gimpel, Sharma, and
  Soricut}]{lan2019albert}
Zhenzhong Lan, Mingda Chen, Sebastian Goodman, Kevin Gimpel, Piyush Sharma, and
  Radu Soricut. 2020.
\newblock Albert: A lite bert for self-supervised learning of language
  representations.
\newblock In \emph{{ICLR} 2020, The 5th International Conference on Learning
  Representations}.

\bibitem[{Leys et~al.(2013)Leys, Ley, Klein, Bernard, and
  Licata}]{leys2013detecting}
Christophe Leys, Christophe Ley, Olivier Klein, Philippe Bernard, and Laurent
  Licata. 2013.
\newblock Detecting outliers: Do not use standard deviation around the mean,
  use absolute deviation around the median.
\newblock \emph{Journal of Experimental Social Psychology}, 49(4):764--766.

\bibitem[{Li et~al.(2019)Li, Weston, and Roller}]{li2019acute}
Margaret Li, Jason Weston, and Stephen Roller. 2019.
\newblock Acute-eval: Improved dialogue evaluation with optimized questions and
  multi-turn comparisons.
\newblock \emph{arXiv preprint arXiv:1909.03087}.

\bibitem[{Li et~al.(2017)Li, Su, Shen, Li, Cao, and Niu}]{li2017dailydialog}
Yanran Li, Hui Su, Xiaoyu Shen, Wenjie Li, Ziqiang Cao, and Shuzi Niu. 2017.
\newblock Dailydialog: A manually labelled multi-turn dialogue dataset.
\newblock In \emph{{IJCNLP} 2017, The 8th International Joint Conference on
  Natural Language Processing}, volume~1, pages 986--995.

\bibitem[{Liu et~al.(2016)Liu, Lowe, Serban, Noseworthy, Charlin, and
  Pineau}]{liu2016not}
Chia-Wei Liu, Ryan Lowe, Iulian Serban, Mike Noseworthy, Laurent Charlin, and
  Joelle Pineau. 2016.
\newblock How not to evaluate your dialogue system: An empirical study of
  unsupervised evaluation metrics for dialogue response generation.
\newblock In \emph{{EMNLP} 2016, The 2016 Conference on Empirical Methods in
  Natural Language Processing}, pages 2122--2132.

\bibitem[{Liu et~al.(2019)Liu, Ott, Goyal, Du, Joshi, Chen, Levy, Lewis,
  Zettlemoyer, and Stoyanov}]{liu2019roberta}
Yinhan Liu, Myle Ott, Naman Goyal, Jingfei Du, Mandar Joshi, Danqi Chen, Omer
  Levy, Mike Lewis, Luke Zettlemoyer, and Veselin Stoyanov. 2019.
\newblock Roberta: A robustly optimized bert pretraining approach.
\newblock \emph{arXiv preprint arXiv:1907.11692}.

\bibitem[{Lowe et~al.(2017)Lowe, Noseworthy, Serban, Angelard-Gontier, Bengio,
  and Pineau}]{lowe2017towards}
Ryan Lowe, Michael Noseworthy, Iulian~Vlad Serban, Nicolas Angelard-Gontier,
  Yoshua Bengio, and Joelle Pineau. 2017.
\newblock Towards an automatic turing test: Learning to evaluate dialogue
  responses.
\newblock In \emph{{ACL} 2017, The 55th Annual Meeting of the Association for
  Computational Linguistics}, volume~1, pages 1116--1126.

\bibitem[{Sai et~al.(2019)Sai, Gupta, Khapra, and Srinivasan}]{sai2019reeval}
Ananya~B. Sai, Mithun~Das Gupta, Mitesh~M. Khapra, and Mukundhan Srinivasan.
  2019.
\newblock Re-evaluating adem: A deeper look at scoring dialogue responses.
\newblock In \emph{{AAAI} 2019, The 33rd AAAI Conference on Artificial
  Intelligence}, pages 6220--6227.

\bibitem[{Serban et~al.(2016)Serban, Sordoni, Bengio, Courville, and
  Pineau}]{serban2016building}
Iulian~Vlad Serban, Alessandro Sordoni, Yoshua Bengio, Aaron~C Courville, and
  Joelle Pineau. 2016.
\newblock Building end-to-end dialogue systems using generative hierarchical
  neural network models.
\newblock In \emph{The 30th {AAAI} Conference on Artificial Intelligence},
  volume~16, pages 3776--3784.

\bibitem[{Serban et~al.(2017)Serban, Sordoni, Lowe, Charlin, Pineau, Courville,
  and Bengio}]{serban2017hierarchical}
Iulian~Vlad Serban, Alessandro Sordoni, Ryan Lowe, Laurent Charlin, Joelle
  Pineau, Aaron~C. Courville, and Yoshua Bengio. 2017.
\newblock A hierarchical latent variable encoder-decoder model for generating
  dialogues.
\newblock In \emph{{AAAI} 2017, The 31st {AAAI} Conference on Artificial
  Intelligence}, pages 3295--3301.

\bibitem[{Shimanaka et~al.(2019)Shimanaka, Kajiwara, and
  Komachi}]{shimanaka2019machine}
Hiroki Shimanaka, Tomoyuki Kajiwara, and Mamoru Komachi. 2019.
\newblock Machine translation evaluation with bert regressor.
\newblock \emph{arXiv preprint arXiv:1907.12679}.

\bibitem[{Sutskever et~al.(2014)Sutskever, Vinyals, and
  Le}]{sutskever2014sequence}
Ilya Sutskever, Oriol Vinyals, and Quoc~V Le. 2014.
\newblock Sequence to sequence learning with neural networks.
\newblock In \emph{{NIPS} 2014, Advances in Neural Information Processing
  Systems 27: Annual Conference on Neural Information Processing Systems 2014},
  pages 3104--3112.

\bibitem[{Tao et~al.(2018)Tao, Mou, Zhao, and Yan}]{tao2018ruber}
Chongyang Tao, Lili Mou, Dongyan Zhao, and Rui Yan. 2018.
\newblock Ruber: An unsupervised method for automatic evaluation of open-domain
  dialog systems.
\newblock In \emph{{AAAI} 2018, The 32nd AAAI Conference on Artificial
  Intelligence}.

\bibitem[{Wolf et~al.(2019)Wolf, Sanh, Chaumond, and
  Delangue}]{wolf2019transfertransfo}
Thomas Wolf, Victor Sanh, Julien Chaumond, and Clement Delangue. 2019.
\newblock Transfertransfo: A transfer learning approach for neural network
  based conversational agents.
\newblock \emph{arXiv preprint arXiv:1901.08149}.

\bibitem[{Zhang et~al.(2018)Zhang, Dinan, Urbanek, Szlam, Kiela, and
  Weston}]{zhang2018personalizing}
Saizheng Zhang, Emily Dinan, Jack Urbanek, Arthur Szlam, Douwe Kiela, and Jason
  Weston. 2018.
\newblock Personalizing dialogue agents: I have a dog, do you have pets too?
\newblock In \emph{{ACL} 2018, The 56th Annual Meeting of the Association for
  Computational Linguistics}, volume~1, pages 2204--2213.

\bibitem[{Zhao and Kawahara(2019)}]{zhao2019effective}
Tianyu Zhao and Tatsuya Kawahara. 2019.
\newblock Effective incorporation of speaker information in utterance encoding
  in dialog.
\newblock \emph{arXiv preprint arXiv:1907.05599}.

\end{thebibliography}
\bibliographystyle{acl_natbib}

\begin{table*}[!t]
    \centering
    \begin{tabu} to 1.0\columnwidth {c|c}
        \textbf{Krippendorff's $\alpha$} & \textbf{Interpretation} \\
        \hline
        \hline
        $<$0.67 & not good \\
        0.67$\sim$0.8 & allowing tentative conclusions to be drawn \\
        $>$0.8 & good reliability
    \end{tabu}
    \caption{Interpretation of Krippendorff's $\alpha$. (\S\ref{sec:agreement})}
    \label{tab:krippendorff}
\end{table*}

\begin{table*}[t!]
    \centering
    \begin{tabu} to 1.0\columnwidth {c|c|c|c}
        \textbf{Hyper-parameter} & \textbf{ADEM} & \textbf{RUBER} & \textbf{RoBERTa-eval} \\
        \hline
        \hline
        \multicolumn{4}{c}{Unsupervised Training} \\
        \hline
        learning rate & & 1e-4 & 3e-6 \\
        batch size & & 30 & 3 \\
        epochs & & 30 & 2 \\
        \hline
        \multicolumn{4}{c}{Supervised Training} \\
        \hline
        learning rate & 1e-3 & 1e-4 & 3e-6 \\
        batch size & 30 & 30 & 3 \\
        epochs & 50 & 50 & 50 \\
    \end{tabu}
    \caption{Optimization hyper-parameters.}
    \label{tab:opt_param}
\end{table*}

\begin{figure*}[!h]
    \centering
    \begin{subfigure}{.24\textwidth}
        \centering
        \includegraphics[width=1.0\linewidth]{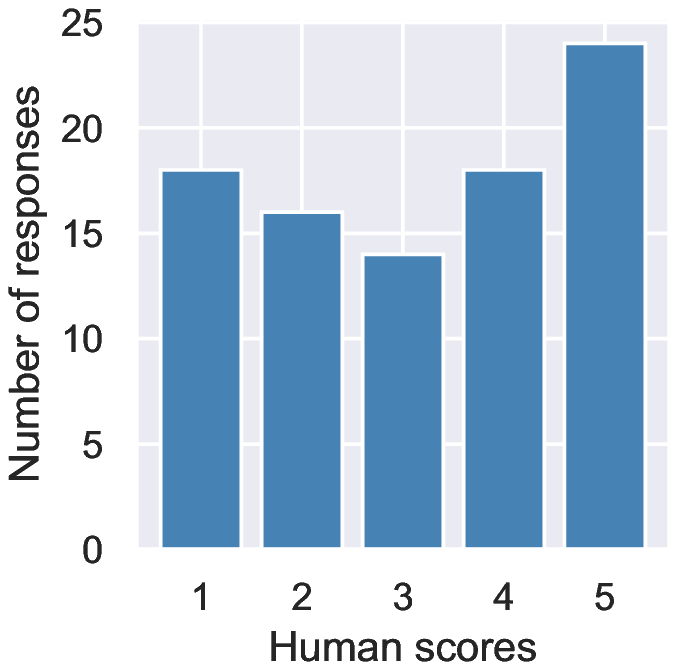}  
        \caption{Human}
    \end{subfigure}
    \begin{subfigure}{.24\textwidth}
        \centering
        \includegraphics[width=1.0\linewidth]{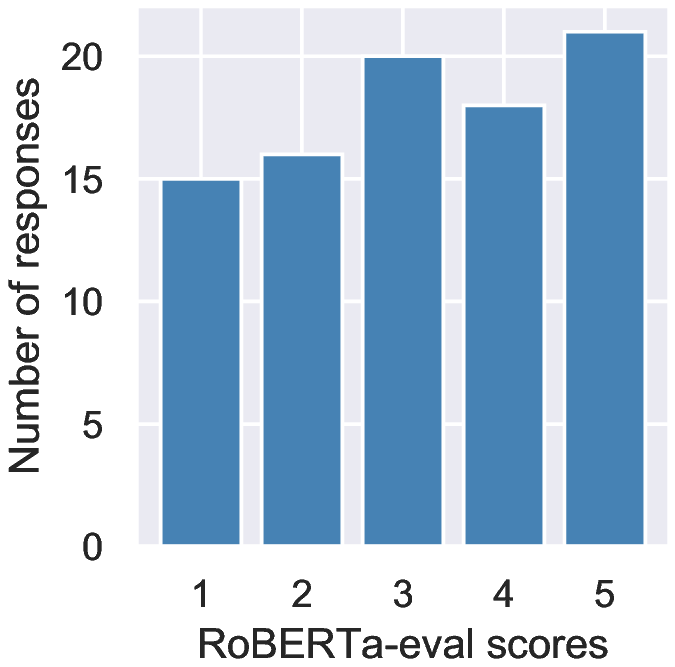}  
        \caption{RoBERTa-eval}
    \end{subfigure}
    \begin{subfigure}{.24\textwidth}
        \centering
        \includegraphics[width=1.0\linewidth]{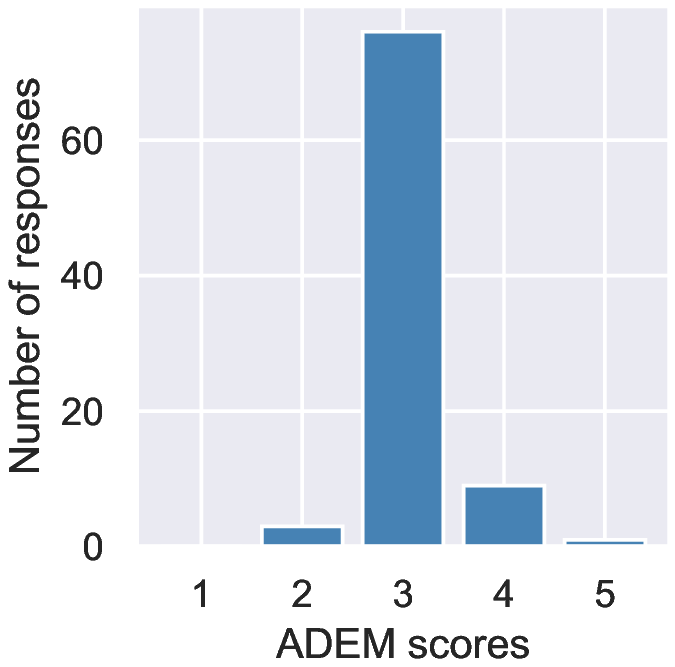}  
        \caption{ADEM}
    \end{subfigure}
    \begin{subfigure}{.24\textwidth}
        \centering
        \includegraphics[width=1.0\linewidth]{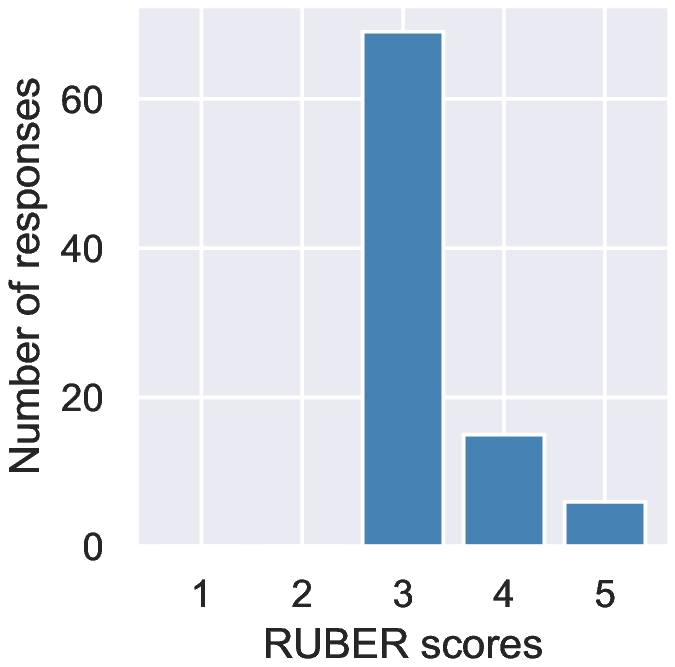}  
        \caption{RUBER}
    \end{subfigure}
\caption{Distributions of human annotations and model outputs on the test data (90 responses). }
\label{fig:test_score_dist}
\end{figure*}

\appendix

\section{Inter-annotator Agreement and Outlier Removal}
\label{sec:agreement}
In the process of collecting human annotations (\S\ref{sec:data}), we collect 3,600 scores in total from 185 Amazon MTurk workers (4 scores for each context-response pair). To assess the data's reliability, we use the Krippendorff's $\alpha$~\citep{krippendorff2018content} instead of commonly used Cohen's $\kappa$ and Fleiss' $\kappa$, because Krippendorff's $\alpha$ can handle 1) an arbitrary number of annotators, 2) various levels of measurement (e.g. nominal, interval), and 3) missing data.

The Krippendorff's $\alpha$ of the original 3,600 annotations of response \textit{appropriateness} is 0.431, which is considered not good according to the interpretation of the number in Table~\ref{tab:krippendorff}. Therefore, we decided to remove the outliers to improve the inter-annotator agreement. We detected outliers for each of the 900 four-annotation groups using the median absolute deviation (MAD) method~\citep{leys2013detecting}. By setting the deviation threshold as 1.0, we identified 895 annotations as outliers. On the remaining 2,705 annotations (roughly 1 annotation is removed for each group), the Krippendorff's $\alpha$ reaches 0.815, which suggests that the data is reliable for the subsequent experiments.

\section{Experimental Settings}
The ADEM and RUBER models use a 2-layer bidirectional gated recurrent unit (BiGRU) sentence encoder with 500 hidden units and a 2-layer BiGRU dialogue encoder with 500 hidden units. The encoders are initialized with the parameters of a pretrained HRED's encoders of the same architecture. To encode speaker information, we concatenate each sentence embedding with a 30-dimensional speaker embedding that indicates whether the sentence's speaker is identical to the response's speaker~\citep{zhao2019effective}. Principal component analysis (PCA) is applied to project response and context embeddings into low-dimensional vectors in ADEM. The number of principal components is 50. The RoBERTa evaluator is based on a pretrained RoBERTa\textit{-large} model, and we finetune the entire model in our experiments. 

Table~\ref{tab:opt_param} shows the hyper-parameters in unsupervised training and supervised training. Following the original paper, we freeze the ADEM's encoders and only finetune its parameters $M$ and $N$, and thus a larger learning rate is used for ADEM. In all experiments, we decay the learning rate with a 0.1 decay rate when a model's validation loss does not improve and stop training early if the learning rate is less than 1e-7.

\section{Model Output Distributions}
The distribution of human annotation scores on the 900 annotated responses has been given in Figure~\ref{fig:anno_score_dist}. To analyze the distribution of model outputs, we show the distributions of human annotation, ADEM's outputs, RUBER's outputs, and RoBERTa-eval's outputs on the test data of 90 responses in Figure~\ref{fig:test_score_dist}. We found that: 1) The distribution of human score is similar to that in Figure~\ref{fig:anno_score_dist}. 2) The proposed RoBERTa evaluator's output has a flatter distribution than human scores. 3) The baseline RUBER and ADEM both have very peaky pseudo-Gaussian distributions whose means are around 3.

\section{Robustness to Changes in Input and Output}
We conduct two sets of experiments to see whether the RoBERTa evaluator's performance would be affected by a slight change in its input and output.

\textbf{Adding Gaussian Noise to Input}. We added Gaussian noise ($\mu$ = 0.0) to human annotations and ran 100 trials with random seeds from 1 to 100. With $\sigma$ = 0.1, the RoBERTa evaluator's performance doesn't change much (Pearson's correlation from 0.64 to 0.64, Spearman's correlation from 0.66 to 0.65). With $\sigma$ = 0.5, the performance degrades more (Pearson's correlation from 0.64 to 0.61, Spearman's correlation from 0.66 to 0.62). Considering that 0.5 $\sigma$ is high and may skew the original human judgement, we believe the evaluator is not greatly affected by the noise.

\textbf{Discretizing Output}. We also tried discretizing the evaluator's outputs (from [1, 5] to \{1, 2, 3, 4, 5\}) and observed a minimal improvement (Pearson's correlation from 0.64 to 0.65, Spearman's correlation from 0.66 to 0.66). Generally speaking, there is no dramatic change in the model's performance when we apply these transformations to the output scores. We believe this shows our model to be fairly robust.

\end{document}